\documentclass{article}
\usepackage{ijcai20}

\usepackage{times}
\usepackage{soul}
\usepackage{url}
\usepackage[hidelinks]{hyperref}
\usepackage[utf8]{inputenc}
\usepackage[small]{caption}
\usepackage{graphicx}
\usepackage{amsmath}
\usepackage{amsthm}
\usepackage{booktabs}
\usepackage{algorithm}
\usepackage{algorithmic}
\title{Towards Question Format Independent Numerical Reasoning: A Set of Prerequisite Tasks }

\author{
Swaroop Mishra$^1$
\and
Arindam Mitra$^2$\and
Neeraj Varshney$^1$\and
Bhavdeep Sachdeva$^1$\And
Chitta Baral$^1$
\affiliations
$^1$Arizona State University\\
$^2$Microsoft Research\\
\emails
\{srmishr1, nvarshn2, bssachde, chitta\}@asu.edu,
arindam.mitra@microsoft.com,
}

\begin{document}

\maketitle

\begin{abstract}

Numerical reasoning is often important to accurately understand the world. Recently, several format-specific datasets have been proposed, such as numerical reasoning in the settings of Natural Language Inference (NLI), Reading Comprehension (RC), and Question Answering (QA). Several format-specific models and architectures in response to those datasets have also been proposed. However, there exists  a strong need for a benchmark which can evaluate the abilities of models, in performing question format independent numerical reasoning, as (i) the numerical reasoning capabilities we want to teach are not controlled by question formats, (ii) for numerical reasoning technology to have the best possible application, it must be able to process language and reason in a way that is not exclusive to a single format, task, dataset or domain. In pursuit of this goal, we introduce NUMBERGAME, a multifaceted benchmark to evaluate model performance across numerical reasoning tasks of eight diverse formats. We add four existing question types in our compilation. Two of the new types we add are about questions that require external numerical knowledge, commonsense knowledge and domain knowledge. 
While recently many QA datasets involving external knowledge have been proposed, ours incorporates them in a numerical reasoning setting. 
Other types in our compilation build upon existing data sets. 
For building a more practical numerical reasoning system, NUMBERGAME demands four capabilities  beyond numerical reasoning: (i) detecting question format directly from data (ii) finding intermediate common format to which every format can be converted (iii) incorporating commonsense knowledge (iv)  handling data imbalance across formats. We build several baselines, including a new model based on knowledge hunting using a cheatsheet. However, all baselines perform poorly in contrast to the human baselines, indicating the hardness of our benchmark. Our work takes forward the recent progress in generic system development, demonstrating the scope of these under-explored tasks.

\end{abstract}

\section{Introduction}
Neural language models powered by datadriven approaches have achieved human level performance across several NLP tasks. However, we still require machines that understand the world well enough to perform reasoning. This capability would give rise to new opportunities for real-world applications such as education, medicine, and scientific discovery \cite{clark2016my} \cite{clark2015elementary}. Additionally, numbers help us to reason in everyday tasks ranging from buying vegetables to reading newspaper to understanding economic situation, survey results, sports, climate change, and election. Since numbers make our conversation accurate, the skill to reason with them is of primary importance in understanding natural language. \cite{dehaene2011number,ravichander2019equate,frank2008number}.


Several datasets have been proposed to foster research in numerical reasoning in natural language understanding and QA context. Examples include DROP \cite{dua2019drop}, EQUATE \cite{ravichander2019equate}, MathQA \cite{amini2019mathqa} and the Mathematics dataset \cite{saxton2019analysing}. But each of these focuses on a specific format. For example, DROP is in RC setting. EQUATE and MathQA are in NLI and QA setting respectively. Since numerical reasoning is generic and independent of any particular format, there exists  a strong need for a benchmark which can evaluate abilities of models in performing question format independent numerical reasoning. Additionally, for numerical reasoning technology to have the best possible application, it must be able to process language and reason in a way that is not exclusive to a single format, task, dataset or domain. In pursuit of this goal, we introduce NUMBERGAME, a multifaceted dataset that builds upon existing datasets where we add some new types. Such a compilation by including datasets with limited training data is intended to encourage the development of general models that can address diverse question formats. For building a more practical numerical reasoning system, NUMBERGAME demands four additional capabilities  beyond numerical reasoning (i) detect question format directly from data (ii) find intermediate common format to which every format can be converted (iii) incorporate commonsense knowledge (iv)  handle data imbalance across formats.

It is now well understood that natural language understanding requires knowledge beyond that is present in the specific text one is trying to understand. This was emphasized when the Winograd schema challenge was proposed \cite{levesque2012winograd} and Marcus \& Davis further emphasize this in their recent book \cite{marcus2019rebooting}. Indeed, several QA datasets have recently been proposed where answering requires reasoning with external knowledge of various kinds such as domain knowledge and commonsense knowledge \cite{bisk2019piqa,marino2019ok}. In our compilation, we add a significant number of QA pairs where external knowledge is needed to answer questions in numerical reasoning setting.



Even though Neural language models such as GPT \cite{radford2018improving}, BERT \cite{devlin2018bert} and RoBERTa (\cite{liu2019roberta} have become standard tools across several language understanding tasks, they can not perform complex forms of reasoning, specifically numerical reasoning \cite{wallace2019nlp}. 
Recently, several neuro-symbolic models have been proposed in response to the numerical reasoning  datasets. Examples include NumNet+v2 \cite{ran2019numnet}, BERT Calculator \cite{andor2019giving} and tag-based multi-span extraction model \cite{efrat2019tag}. Though they have performed well on the DROP dataset, they are limited in terms of the type of questions and numerical operations they handle. Among those models, we selected NumNet+v2, the best one for which code is publicly available. We added a question type converter module on top of NumNet+v2 and created an initial baseline for our dataset. Noting that this model does not take into account external knowledge, we created an new enhanced architecture that first uses knowledge hunting (searching for the missing knowledge) \cite{banerjee2019careful,mitra2019exploring} with respect to a cheat sheet to identify the needed knowledge. This is inspired by the observation that cheat sheet makes the task easier for humans while solving math questions of various types. We then use this knowledge in the NumNet+V2 setting. This leads to an improved baseline. 


Our contribution in this paper is as follows:

\begin{itemize}
    \item We compile a multifaceted dataset involving eight different types and define a task to solve this in multi-task setting. In the process, we add four new types of data.
    \item We introduce questions that require external numerical knowledge, commonsense knowledge and domain knowledge in the QA setting.
    \item We create a baseline by extending the NumNet+v2 model to handle questions of various types together.
    \item We develop a model that uses knowledge hunting using a cheat sheet together with NumNet+V2; this leads to an improved baseline.
\end{itemize}

\section{Related Works}
\textbf{Datasets for Numerical reasoning}:
Quantitative reasoning has been a challenging problem for a long time. Small question answering datasets were proposed to understand the quantitative aspect of natural language such as the template-based dataset which solved questions with equations as parameters \cite{kushman2014learning}, addition-subtraction dataset \cite{Hosseini14learningto} and arithmetic problems dataset \cite{koncel2015parsing}. Difficulty of questions were increased in subsequent datasets \cite{roy2016solving}, \cite{upadhyay2016learning}. Later, larger datasets were created to facilitate deep learning research \cite{ling2017program}. One of our focus in creating this dataset is to have simple question answering problems and minimize data repetition.
 \\\\
\textbf{Neurosymbolic Models}: NAQANet is the first neuro symbolic model proposed to solve the DROP dataset. It is a numerically-aware QANet model,
which allows the reading comprehension system to generate three new answer types present in DROP.
BERT Calculator uses the BERT embedding by separating the computation part from the neural network and using a calculator tool to do that. Tag-based multi-span extraction model introduces a novel approach to better understand multi-span questions. NumNet has a numerically-aware graph neural network which considers comparison information over numbers in the question and passage. We develop our model on top of NumNet+v2 which is the combination of NumNet+ and tag based multi-span extraction model.
\\\\
\textbf{Knowledge Retrieval}:
Elasticsearch has been shown to be successful in prior works for knowledge retrieval \cite{khot2019s,banerjee2019careful,mitra2019exploring}. We use elasticsearch and a heuristic ranking algorithm for extracting relevant information.\\\\
\textbf{Multi-tasking Benchmarks}:
Several tasks in the BAbI dataset \cite{weston2015towards} were designed to act as prerequisites for any system that aims to be capable of interacting and conversing with a human. GLUE \cite{wang2018glue}, a benchmark to evaluate the performance of models across a diverse set of NLU tasks, had the objective to favor and encourage models that share general linguistic knowledge across tasks. SuperGLUE \cite{wang2019superglue} consists of more difficult language understanding tasks than the GLUE benchmark. ORB \cite{dua2019orb} has an evaluation server which reports performance on seven diverse reading comprehension datasets, encouraging development of a single model for a wide variety of reading phenomena. QuAIL \cite{downeygetting} covers four domains (fiction, blogs, political news, and user story texts), demanding a system to handle both general and text-specific questions which are impossible to answer from pretraining data. DecaNLP \cite{McCann2018decaNLP} poses a challenge that spans ten tasks in multitask setting. It also introduces the Multitask Question Answering Network (MQAN) which jointly learns all tasks in decaNLP without any task-specific modules or parameters in the multitask setting. T5 \cite{raffel2019exploring} introduces a powerful mode that converts every language problem into a text-to-text format, taking forward the transfer learning paradigm. UnifiedQA \cite{khashabi2020unifiedqa} is a recent interesting work where the latest advances in language modeling are used to build a single pre-trained QA model. Our work, NUMBERGAME focuses on prerequiste tasks necessary to do question format independent numerical reasoning. It also demands detection of numerical reasoning question format directly from data and incorporation of commonsense knowledge, which are different from existing works.
\section{Data Creation}\label{createdata}
Our dataset consists of a variety of question types that involve numerical computation and reasoning. 
 We carefully select eight setting that involve numerical reasoning such that each setting has distinct properties and one is not just an extension of others. We divide our dataset into two broad categories. First category includes novel datasets and second is a collection of existing datasets.

\subsection{Novel Datasets}
In this section, we describe datasets that we have created manually.
Recently, a few datasets have been proposed in various areas like Visual Question Answering, Textual Question Answering that require external knowledge to answer the questions. 
Here, we create data in four different setting. Three of them require knowledge, specifically numerical common sense knowledge and the fourth one is a collection of completion based questions.

\paragraph{Daily Life Maths Requiring Numerical Common Sense Knowledge:} This type includes questions that require some common sense numerical knowledge that is not explicitly provided in the question. Table \ref{tab:mising_numerical_knowledge} shows some examples of this category. This dataset creation process consists of two phases. First, we create a list of concepts that involve some common sense numerical knowledge such as 'A dice has 6 faces', 'English language has 5 vowels'. Second, we form questions that leverage this numerical knowledge in various real world contexts. Using this two step approach, we create a total of 404 question and answer pairs.

\begin{table}
    \centering
    \begin{tabular}{ l l r }
    \toprule
        \textbf{Question} &
        \textbf{Knowledge Required} &
        \textbf{Answer}\\
    \midrule
    %
    \multicolumn{1}{p{3.4cm}} {Ella and Lily are playing a game that requires 10 die. Find out the total number of faces in 10 die.}& \multicolumn{1}{p{2.1cm}}{A die has 6 faces}& 60 \\
    \midrule
    \multicolumn{1}{p{3.4cm}}{Jacob and Lillian are running a km long race. Jacob finished the race when Lillian was 190 meters from the finish line. How many meters did Lillian cover till that time?}& \multicolumn{1}{p{2.1cm}}{ 1000 meters make a km } & 810\\
    \midrule
    \multicolumn{1}{p{3.4cm}}{A man can lift one box in each of his hands. How many boxes can a group of 5 people hold in total?}& \multicolumn{1}{p{2.1cm}}{ A human being has 2 hands } & 10 \\
    
    
    \bottomrule
    \end{tabular}
    \caption{Example questions where numerical knowledge required to answer is not explicitly provided in the question.}
    \label{tab:mising_numerical_knowledge}
\end{table}
\paragraph{Application of Maths requiring domain knowledge:}
This dataset includes problems which need the usage of rule-like numeric domain knowledge such as formulae. This sets a higher bar on the ability of a model to use numeric knowledge.  However, language models which have succeeded in most question answering datasets have comparatively less accuracy on the AI2 Reasoning Challenge (ARC) which consists of  grade-school level multiple-choice science questions \cite{clark2018think}. 
In this dataset, we add a significant number of questions belonging to few simple concepts. This can help in assessing the capability of language models to solve science questions involving a small number of concepts.
However, our dataset creation framework can be easily extended to create questions from a larger set of concepts.
We create chemistry questions covering the concepts of balancing the reactions and molecular weight computation of compounds.
We create such questions from a set of 90 reactions and 53 compounds. From Physics, we create numerical questions involving the speed, distance and time. Table \ref{tab:chem} shows some examples of this category.

\begin{table}[ht]
    \centering
    \begin{tabular}{ l l r }
    \toprule
    \textbf{Question} &\textbf{Knowledge Required} & \textbf{Answer} \\
    \midrule
    \multicolumn{1}{p{3.4cm}}{Find the mass percentage of H in C6H6} & \multicolumn{1}{p{2.1cm}}{Mass of C is 12 units and mass of H is 1 units} & 7.69\\
    \midrule
    \multicolumn{1}{p{3.4cm}}{How many units of H2 are required to react with 2 units of C2H4 to form 2 units of C2H6}
    & 
    \multicolumn{1}{p{2.1cm}}{H2 + C2H4 = C2H6}
    &
    2 \\
    \midrule
    
    \multicolumn{1}{p{3.4cm}}{A car covers 912 meters in 19 seconds. If bike's speed is one fourth of the car. Find the distance covered by the bike in 4 seconds.}
    &
    \multicolumn{1}{p{2.1cm}}{distance travelled = speed * time}
    &
    48 \\
    \bottomrule
    \end{tabular}
    \caption{Example questions where domain knowledge is required to answer a question.}
    \label{tab:chem}
\end{table}

\paragraph{Quantitative Comparison Requiring Common Sense Knowledge along with Numerical Common Sense:}
This type covers questions that involve numerical comparison between two quantities.
QUAREL\cite{tafjord2019quarel} is a dataset involving qualitative relationships and quantities in multiple domains such as science and economics.
We select a subset of Quarel questions that involve numerically comparable quantities, and introduce numbers in place of text representing qualitative relationship. A few examples of original Quarel questions and transformed questions are shown in Table \ref{tab:quarel}.
We accumulate a total of 807 questions of this kind.

\begin{table}[ht]
    \centering
    \begin{tabular}{   l l  }
    \toprule
        \textbf{QUAREL Question} &\textbf{Transformed Question} \\
    \midrule

    \multicolumn{1}{p{4cm}}{A person wants to get shopping done quickly. They know that they can get through the checkout at big store faster than they can at small store. The store they go to to finish quickly is
    \newline
    (A) \textbf{big store} (B) small store}
    & \multicolumn{1}{p{4cm}}{ A person wants to get shopping done quickly. They know that they can get through the checkout at big store in 5 minutes whereas it can take 20 mintues at small store. The store they go to to finish quickly is
    \newline
    (A) \textbf{big store} (B) small store}
    \\
    \midrule
    
    \multicolumn{1}{p{4cm}}{Tina is racing her two dogs. Her greyhound is slim, her rottweiler is heavy. The dog that gets faster more quickly is the
    \newline
    (A) rottweiler (B) \textbf{greyhound}}
    &
    \multicolumn{1}{p{4cm}}{Tina is racing her two dogs. Her greyhound weighs 88 lbs and her rottweiler weighs 79 lbs. The dog that gets faster more quickly is the 
    \newline
    (A) \textbf{rottweiler} (B) greyhound}
    \\
    \midrule
    
    \multicolumn{1}{p{4cm}}{A golf ball has a smaller mass then a baseball. Which item has a weaker gravitational field?
    \newline
    (A) \textbf{golf ball} (B) baseball}
    &
    \multicolumn{1}{p{4cm}}{A golf ball has a mass of 78 grams and a baseball has a mass of 0.159 Kg. Which item has a weaker gravitational field?
    \newline
    (A) \textbf{golf ball} (B) baseball}\\
    
    \bottomrule
    \end{tabular}
    \caption{Examples showing conversion of QUAREL questions to quantitative comparison questions}
    \label{tab:quarel}
\end{table}

\paragraph{Completion type questions}
Completion is a type of question where a blank is required to be filled. 
We create such questions from the Arithmetic Word Problem repository \cite{roy2018mapping}\cite{roy2016solving}\cite{roy2017unit} manually following a two step process. First, we introduce a blank in the question. Then, we reformulate the question such that the blank takes place of the answer. We also create adversarial examples of this type. Table \ref{tab:completion} shows some examples of this kind. 

\begin{table}
    \centering
    \begin{tabular}{ l l  }
    \toprule
    \textbf{Arithmetic Word Problem} &\textbf{Transformed Question} \\
    \midrule

    \multicolumn{1}{p{4cm}}{Joan found 70 seashells on the beach. she gave Sam some of her seashells. She has 27 seashell left. How many seashells did she give to Sam ?
    \textbf{43}}
    & 
    \multicolumn{1}{p{4cm}}{Joan found 70 seashells on the beach . She gave Sam some of her seashells . She has 27 seashells left. She gave $\rule{1cm}{0.15mm}$ seashells to Sam.
    \textbf{43}}\\
    \hline
    \multicolumn{1}{p{4cm}}{Last week Tom had 74 dollars. He washed cars over the weekend and now has 86 dollars. How much money did he make washing cars ?
    \textbf{12}}
    & 
    \multicolumn{1}{p{4cm}}{Last week Tom had 74 dollars.  He washed cars over the weekend and made another 86 dollars. Tom has $\rule{1cm}{0.15mm}$ dollars now .
    \textbf{160}}\\
    \bottomrule
    \end{tabular}
    \caption{Examples showing MAWPS questions and corresponding questions in Completion format }
    \label{tab:completion}
\end{table}


\subsection{Collection of Existing Datasets}
We compile questions from some of the existing datasets that involve numerical reasoning. In order to avoid adding repetitive questions and ensure high quality of out dataset, we filter questions following a five step procedure. First, we remove questions that do not have annotated answers. Second, we remove grammatically incorrect questions. Then, we eliminate problems which have high lexical overlap with rest of the dataset, thus ensuring that our dataset incorporates mostly unique concepts. Next, we rectify type mismatch issues such as, "there are 7.0 students" to "there are 7 students" as the number of students is not a float quantity. Finally, we discard invalid and inaccurate questions. 
We compile a total of four question types from the existing datasets.

\paragraph{Reading Comprehension (RC) with Explicit Math}
This category includes reading comprehension questions which require explicit math to answer. We take DROP as the source here and process the data using the above-mentioned 5 step filtering procedure. In DROP, answer can be a number, a date or a segment of text from the passage. We divide questions in two classes based on the answer type. The first type includes questions having a numerical answer and the second type having text segment in the question passage as the answer. 
We include the first type in this category and second type in the next category. 

\paragraph{Reading Comprehension involving Implicit Math}
 Here, we select reading comprehension questions from DROP where the answer is not a numerical value but some sort of mathematical operation such as counting or sorting is required to answer the question. This category is inspired from the fact that many a times in real world some sort of math is implicitly required to answer a question.

\paragraph {Quantitative NLI}
Natural Language Inference (NLI) or Recognizing Textual Entailment (RTE) has been considered as a benchmark task in natural language understanding. Recently introduced EQUATE dataset has quantitative NLI problems combined from diverse sources. We use EQUATE to add NLI questions to our dataset.

\paragraph{Arithmetic Word Problems}
This dataset is a combination of algebra and arithmetic word problems. We collect problems from the Arithmetic Word Problem repository \cite{koncel2016mawps}, SingleEq \cite{koncel2015parsing} and SimulEq-S \cite{kushman2014learning}. Using Spacy sentence similarity \cite{honnibal2017spacy}, we ensure that questions in this type don't have high overlap with questions in completion type.

\section{Combining all types of data}
\paragraph{Partitioning the dataset:}
In a real world setting, number of problems in each type of data is different. Instead of under-sampling or over-sampling data across types, we decide to keep them disproportionately to mimic the real world setting. 
Each type of data was randomly partitioned into training (70\%), development (10\%) and test (20\%) set by ensuring that there is no data leakage among these splits. In order to ensure this for RC problems, we keep all questions of a passage in only one of the splits. Similarly, in other setting, questions which are very similar to each other are kept in only one of the splits. 
This way, we reduce the possibility of memorization by language models.
\paragraph{Validation:}
The validation of the test set of our novel datasets was performed by three individuals.\footnote{None of the authors were involved in this process}. We provided them questions and asked them to mark questions as either valid or invalid. A very small percentage of the data was marked as invalid which we later filtered out.

\begin{figure*}
\includegraphics[width=\linewidth]{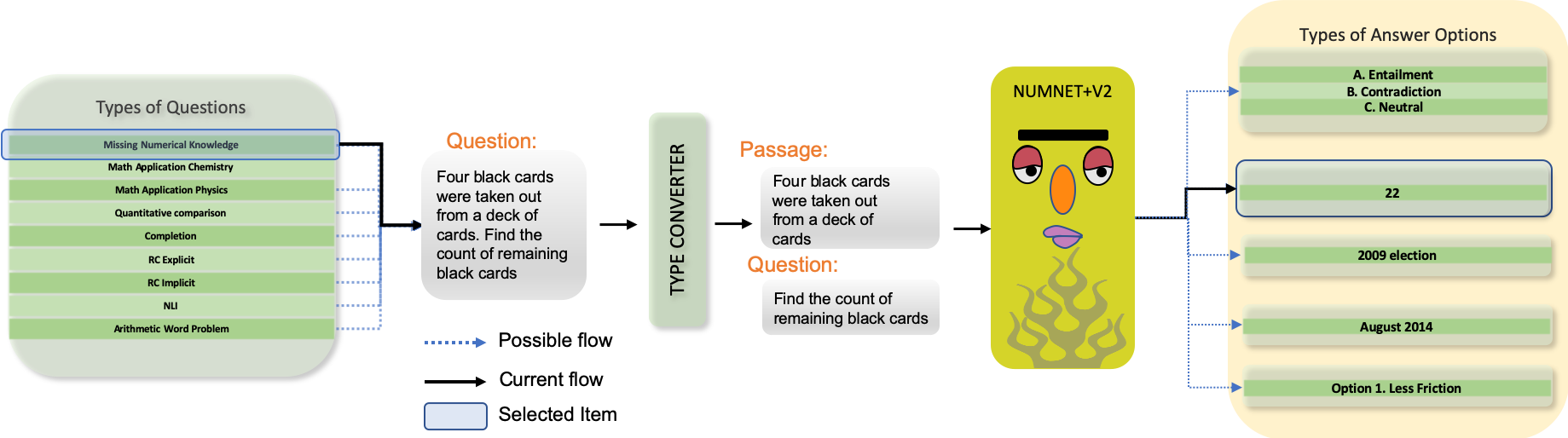}
   \caption{Baseline Model overview depicting the various questions types and their answers.}
\label{fig:motivation}
\end{figure*}
\section{Data Analysis}
We have performed data analysis in order to gauge the quality of our dataset.

\begin{figure}[h]
\centering
\includegraphics[width=8cm, height=4cm]{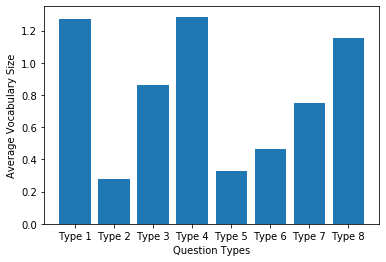}
\caption{Distribution of average vocabulary in the data set.}
\label{VocabDistribution}
\end{figure}

\paragraph{Vocabulary Size:} First, we calculate vocabulary size of each type by finding the number of unique words across all questions. Since our dataset is unbalanced in terms of question type, we find the average vocabulary size of the dataset by dividing vocabulary size with number of data in that type. 
\paragraph{Which data has more average vocabulary size?} As illustrated in Figure \ref{VocabDistribution}. Most of the question types belonging to the novel dataset category have relatively better average vocabulary size. This implies 
questions in those types have less repetitiveness. 


\begin{table}[ht]
    \centering
    \begin{tabular}{llrrr}
    \toprule
    \multicolumn{1}{p{1cm}}{\textbf{Data Type}} &
    \multicolumn{1}{p{3.5cm}}{\textbf{Problem}}  &
    \multicolumn{1}{p{0.8cm}}{\textbf{Train}} &
    \multicolumn{1}{p{0.8cm}}{\textbf{Dev}} &
    \multicolumn{1}{p{0.8cm}}{\textbf{Test}} \\
    \midrule
    \multicolumn{1}{p{1cm}}{Type 1}
    &
    \multicolumn{1}{p{3.5cm}}{Missing Numerical Knowledge}
    &
    \multicolumn{1}{p{0.8cm}}{282}
    &
    \multicolumn{1}{p{0.8cm}}{41}
    &
    \multicolumn{1}{p{0.8cm}}{81}
     \\
    \multicolumn{1}{p{1cm}}{Type 2} 
     &
     \multicolumn{1}{p{3.5cm}}{Maths in other domains}
     &
     \multicolumn{1}{p{0.8cm}}{1131}
     & 
     \multicolumn{1}{p{0.8cm}}{164}	
     &
     \multicolumn{1}{p{0.8cm}}{325} \\
    \multicolumn{1}{p{1cm}}{Type 3} 
    &
    \multicolumn{1}{p{3.5cm}}{Quantitative comparison}
    &
    \multicolumn{1}{p{0.8cm}}{564}	
    & 
    \multicolumn{1}{p{0.8cm}}{81}
    & 
    \multicolumn{1}{p{0.8cm}}{162} \\
    
    \multicolumn{1}{p{1cm}}{Type 4}
    &
    \multicolumn{1}{p{3.5cm}}{Completion Type}
    &
    \multicolumn{1}{p{0.8cm}}{770}	
    & 
    \multicolumn{1}{p{0.8cm}}{110}
    &
    \multicolumn{1}{p{0.8cm}}{220} \\
    \multicolumn{1}{p{1cm}}{Type 5}
    &
    \multicolumn{1}{p{3.5cm}}{Reading comprehension with Explicit Math}
    &
    \multicolumn{1}{p{0.8cm}}{37949}	
    &
    \multicolumn{1}{p{0.8cm}}{5421}
    &
    \multicolumn{1}{p{0.8cm}}{10842} \\

    \multicolumn{1}{p{1cm}}{Type 6}
    &
    \multicolumn{1}{p{3.5cm}}{Reading Comprehension with Implicit Math}
    &
    \multicolumn{1}{p{0.8cm}}{22908}
    &
    \multicolumn{1}{p{0.8cm}}{3272}
    &
    \multicolumn{1}{p{0.8cm}}{6544}
    \\
    \multicolumn{1}{p{1cm}}{Type 7}
    &
    \multicolumn{1}{p{3.5cm}}{Quantitative NLI}
    &
    \multicolumn{1}{p{0.8cm}}{6791}
    &
    \multicolumn{1}{p{0.8cm}}{970}
    &
    \multicolumn{1}{p{0.8cm}}{1941} \\
    
    \multicolumn{1}{p{1cm}}{Type 8}
    &
    \multicolumn{1}{p{3.5cm}}{Arithmetic Word Problems}
    &
    \multicolumn{1}{p{0.8cm}}{886}
    &
    \multicolumn{1}{p{0.8cm}}{126}
    &
    \multicolumn{1}{p{0.8cm}}{254} \\
    \bottomrule

    \end{tabular}
    \caption{Dataset size for all the question types across all splits }
    \label{tab:datasize}
\end{table}

\begin{figure}[h]
\centering
\includegraphics[width=7cm, height=5cm]{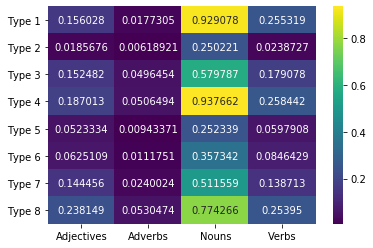}
\caption{Distribution of Part of Speech tags across question types.}
\label{Parts_of_Speech}
\end{figure}

\begin{figure}[h]
\centering
\includegraphics[width=4cm, height=4cm]{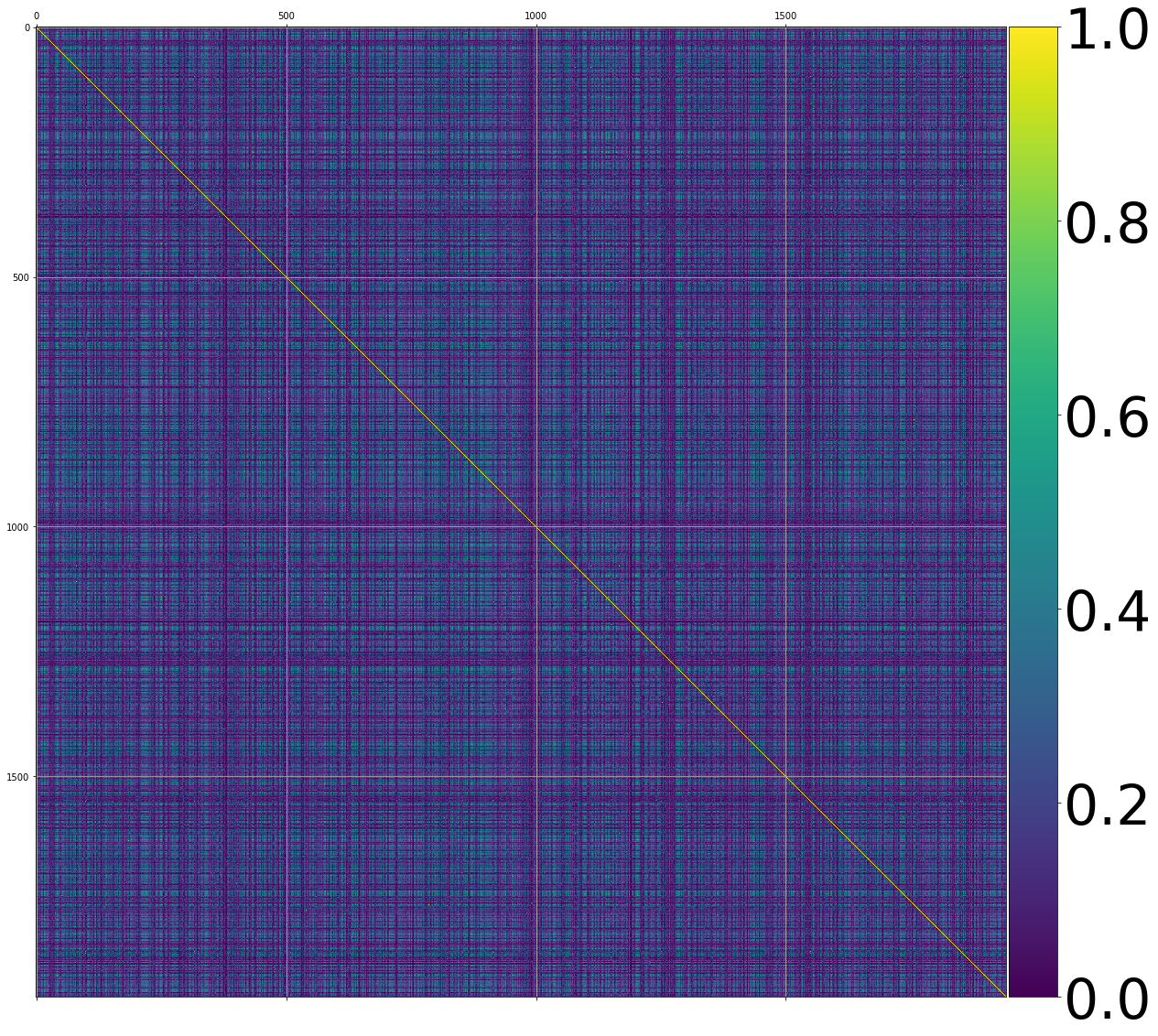}
\includegraphics[width=4cm, height=4cm]{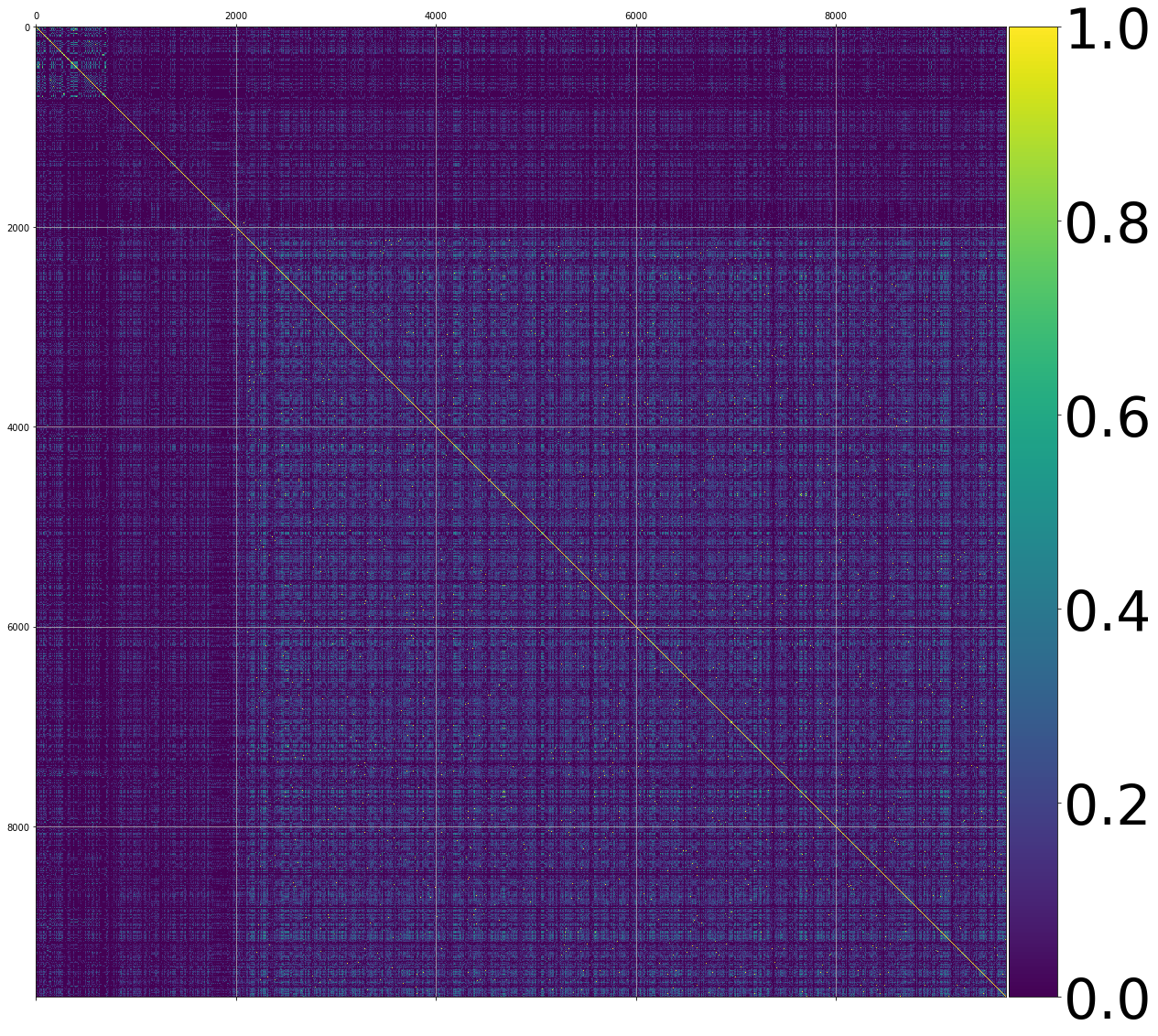}
\caption{Similarity Charts for QUARREL and EQUATE }
\label{similarityChart_1}
\end{figure}

\begin{figure}[h]
\centering
\includegraphics[width=4cm, height=4cm]{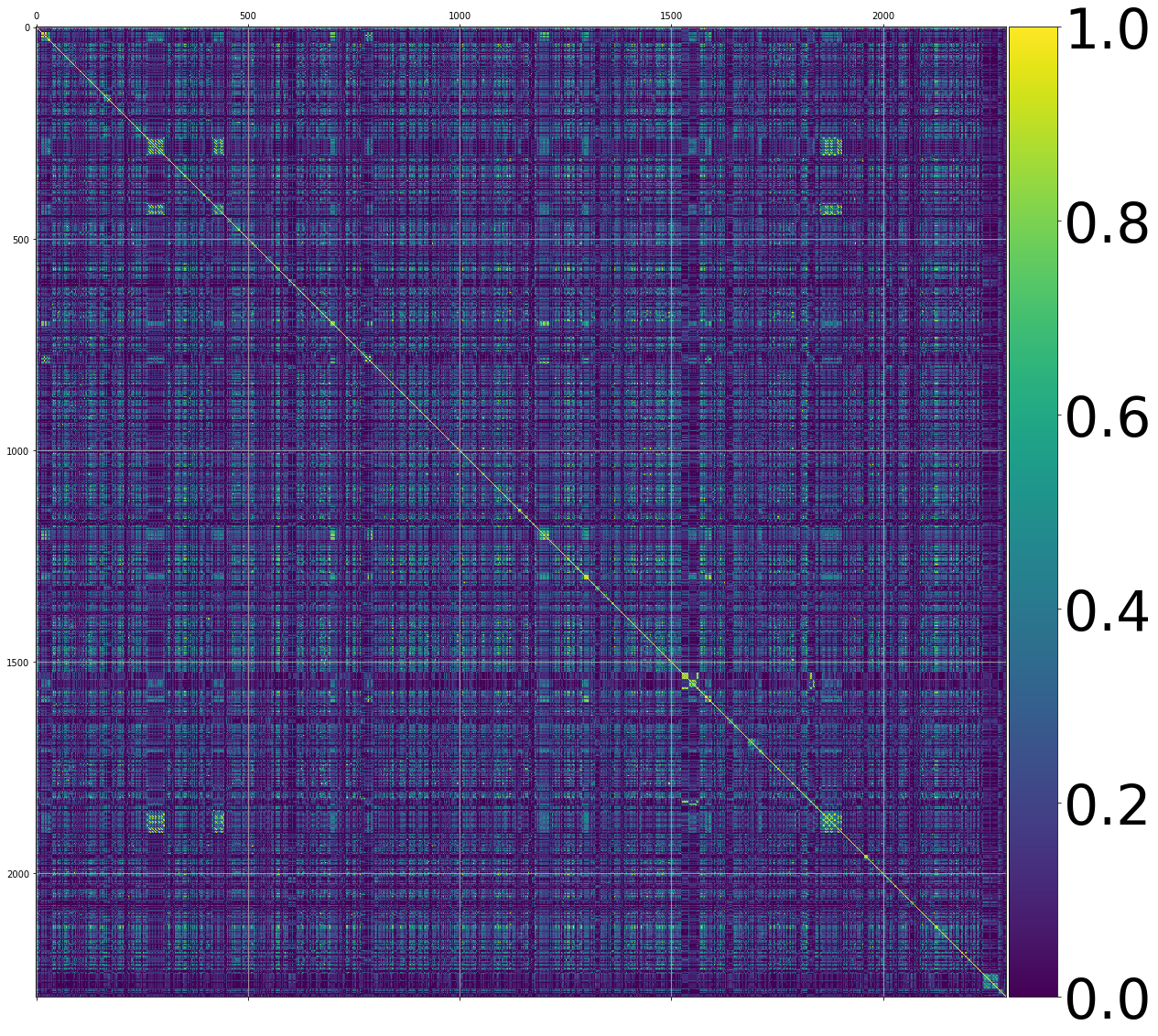}
\includegraphics[width=4cm, height=4cm]{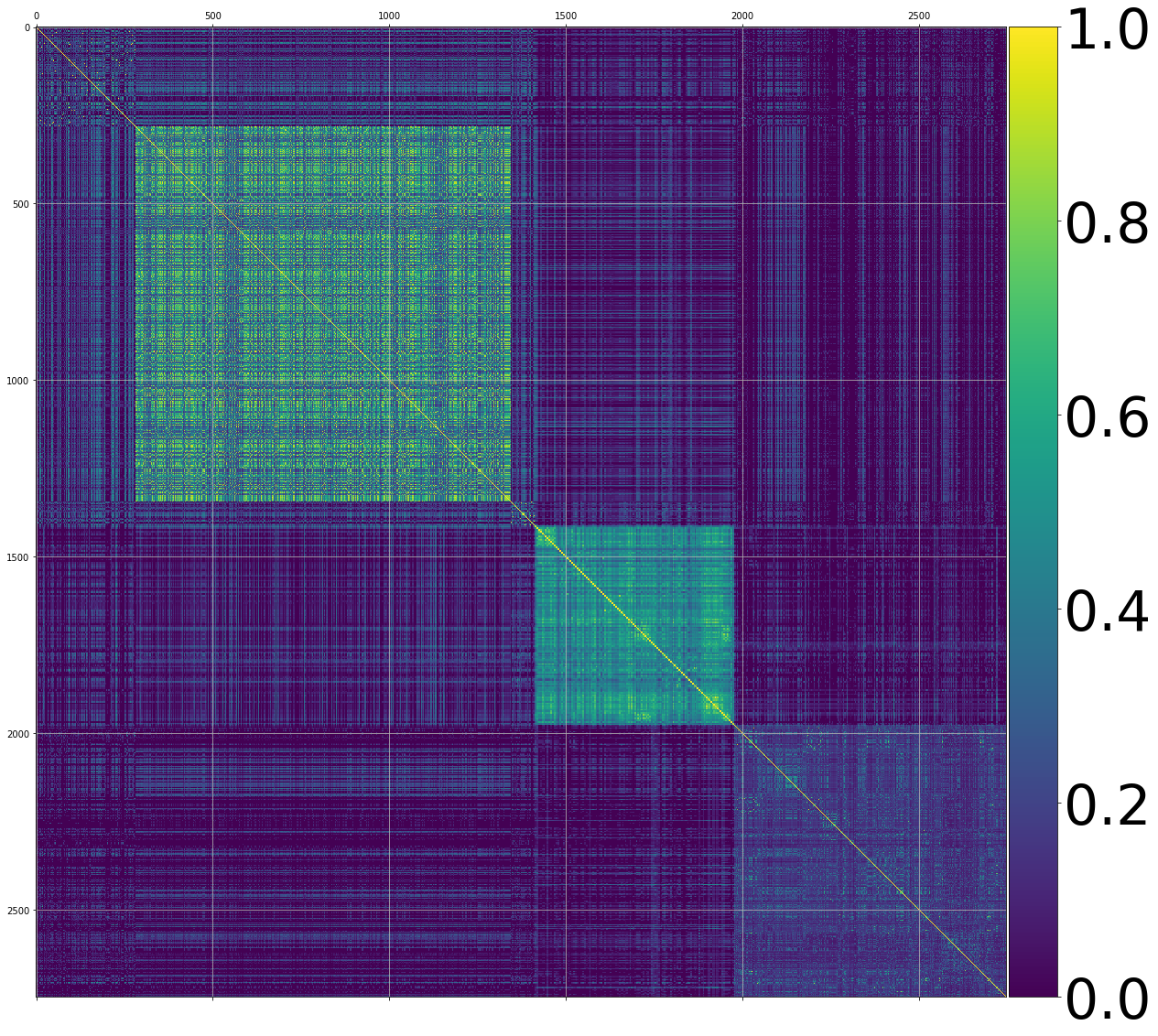}
\caption{Similarity Charts for DROP and our manually created data.}
\label{similarityChart_2}
\end{figure}

We expand on our vocabulary analysis to understand Figure \ref{VocabDistribution} further. We dive deep to analyze different parts of speech. Figure \ref{Parts_of_Speech} summarises our analysis. Most of the novel datasets have more average number of nouns, verbs and adjectives implying there are more types of entities, actions and attributes. This further means that those data-sets are more diverse in nature.
\paragraph{Sentence Similarity Analysis}
We further extend our analysis to reinforce our inference from the word vocabulary analysis. We find cosine similarity of a sentence with every other sentence. Figure \ref{similarityChart_1} and \ref{similarityChart_2} illustrate the similarity analysis for various datasets.

\paragraph{Which data consists of most dissimilar sentences?}
As illustrated in Figure \ref{similarityChart_1}, most question sentences in Quarrel \cite{tafjord2019quarel} have high similarity implying that data is repetitive. Same is true for majority of EQUATE data. From figure \ref{similarityChart_2}, it is evident that DROP also has high similarity among questions. We also analyze this similarity metric for our dataset and find that similarity among questions is significantly less. 
Some similarity boxes can be seen in the plot. They are mostly due to Type 2 data and partly due to Type 3. This implies that our dataset is far less repetitive than others. Also, in our dataset the repetition is sparse and is not equally distributed among the whole dataset unlike others. This way, our dataset is more diverse.

\paragraph{Why Type 2 questions have small vocabulary and high similarity?}
This is because, type 2 consists of math questions in other domains. Most of the questions in this category are chemistry questions in text book setting. Since we are limiting the number of concepts to two and questions in chemistry follow a pattern, this type results in comparatively less vocabulary and high sentence overlap. 

\begin{figure*}
\includegraphics[width=\linewidth]{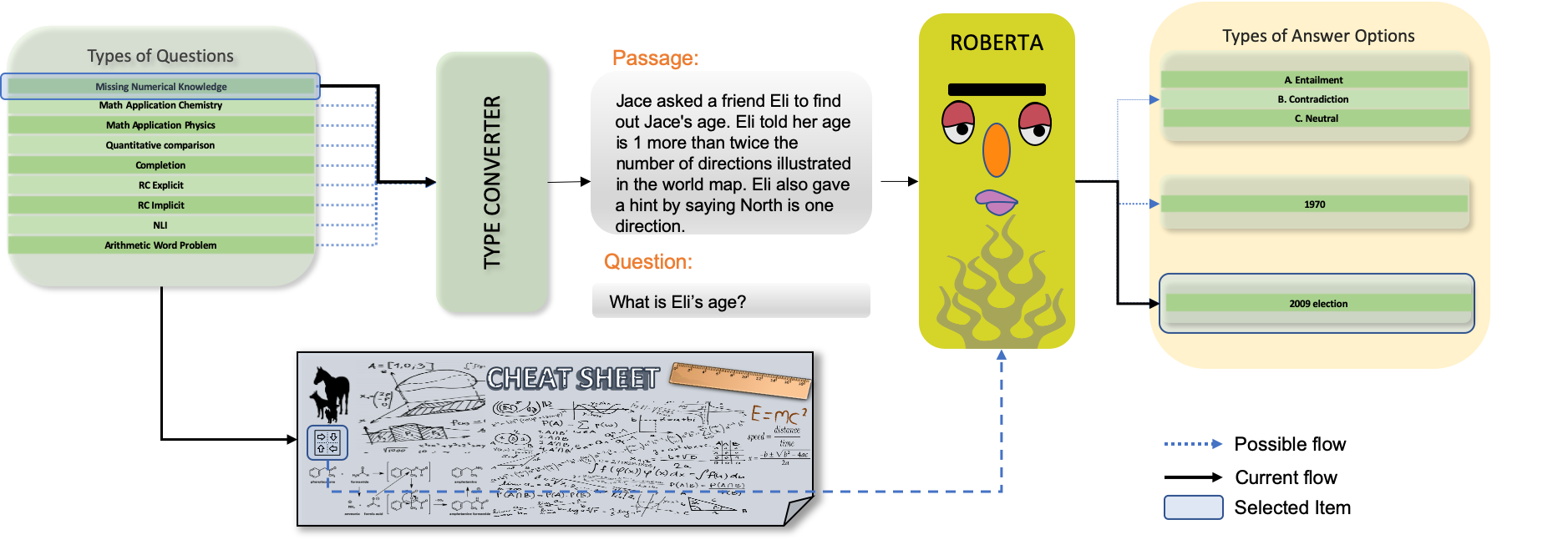}
   \caption{Knowledge Hunting augmented Model overview depicting the various questions types and their answers.}
\label{fig:knowledgemodel}
\end{figure*}
\begin{figure*}
\includegraphics[width=\linewidth]{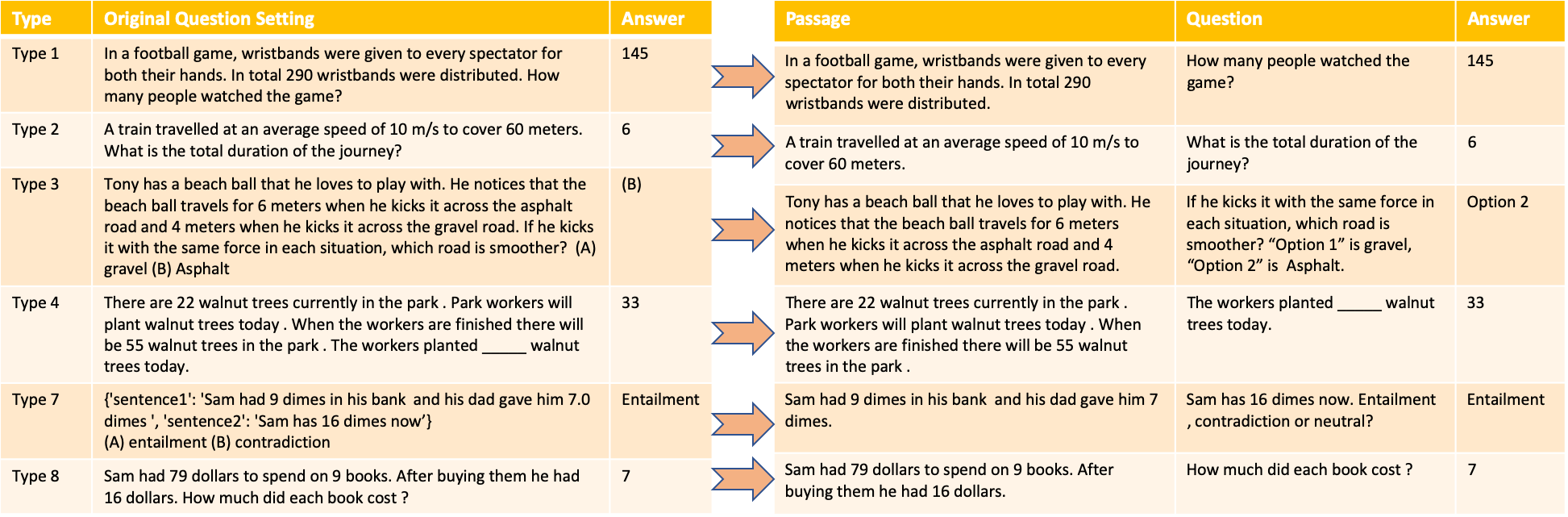}
   \caption{Conversion of various question types to reading comprehension format}
\label{motiv}
\end{figure*}


\begin{table*}[t]
\centering
\scriptsize
\resizebox{2\columnwidth}{!}{
\begin{tabular}{ c c c c c c c c c c c c c c c c c }
\toprule 
                                  & \multicolumn{2}{l|}{Type 1}   & \multicolumn{2}{l|}{Type 2}     & \multicolumn{2}{l|}{Type 3}     & \multicolumn{2}{l|}{Type 4}     & \multicolumn{2}{l|}{Type 5}     & \multicolumn{2}{l|}{Type 6}    & \multicolumn{2}{l|}{Type 7}     & \multicolumn{2}{l}{Type 8}     \\ \midrule
                                  & EM            & F1            & EM             & F1             & EM             & F1             & EM             & F1             & EM             & F1             & EM             & F1            & EM             & F1             & EM             & F1             \\ \midrule
\begin{tabular}[c]{@{}l@{}}Random Baseline\end{tabular}                   & 0             & 0             & 0.31           & 0.31           & 46.9           & 46.9           & 0              & 0              & 0.53           & 0.53           & 1.6            & 3.44          & 33.02          & 33.02          & 0.39           & 0.39           \\
\begin{tabular}[c]{@{}l@{}}Majority Baseline\end{tabular}                 & 1.23          & 1.23          & 13.85          & 13.85          & 50             & 50             & 0.45           & 0.45           & 7.41           & 7.41           & 1.67           & 3.8           & 36.53          & 36.53          & 1.18           & 1.18           \\
\begin{tabular}[c]{@{}l@{}}Question-only Baseline\end{tabular}            & 1.23          & 1.23          & 13.23          & 13.23          & 23.21          & 25.07          & 0.45           & 0.45           & 6.01           & 6.12           & 20.86          & 25.14         & 32.82          & 32.82          & 2.36           & 2.36           \\
\begin{tabular}[c]{@{}l@{}}Context-only Baseline\end{tabular}             & 1.23          & 1.23          & 14.15          & 14.15          & 0              & 22.75          & 19.09          & 19.09          & 0.51           & 0.57           & 0.89           & 3.02          & 0              & 0              & 9.45           & 9.45           \\
\begin{tabular}[c]{@{}l@{}}Extended NumNet+v2\end{tabular}                & 0             & 0             & 37.54          & 37.54          & \textbf{58.02} & \textbf{58.02} & 31.36          & 31.36          & 68.06          & 68.23          & \textbf{57.23} & \textbf{70.2} & 85.73          & 85.73          & \textbf{23.23} & \textbf{23.23} \\
\begin{tabular}[c]{@{}l@{}}Extended NumNet+V2 + Knowledge\end{tabular}    & 4.94          & 5.56          & 37.54          & 37.54          & 46.30          & 46.57          & \textbf{36.36} & \textbf{36.36} & \textbf{68.40} & \textbf{68.59} & 57.20          & 69.61         & \textbf{85.88} & \textbf{85.88} & 22.44          & 22.44          \\
\begin{tabular}[c]{@{}l@{}}Extended NumNet+V2 + Oversampling\end{tabular} & \textbf{7.41} & \textbf{7.41} & \textbf{38.77} & \textbf{38.77} & 47.53          & 47.84          & 35.91          & 35.91          & 43.99          & 44.30          & 40.46          & 53.71         & 85.37          & 85.39          & 22.44          & 22.44          \\
\begin{tabular}[c]{@{}l@{}}Human baseline\end{tabular}                    & 94.42         & 94.42         & 94.5           & 94.5           & 97.75          & 97.75          & 95.0           & 95.0           & 94.5           & 94.67          & 95.0           & 96.1          & 96.5           & 96.5           & 92.75          & 92.75         
\end{tabular}
}
\caption{Table showing the performance of various baselines on our test set across all data types}
\label{results}
\end{table*}
\section{Baseline Models}
We evaluate our dataset using several baselines including heuristic baseline, extended NumNet+v2, bias-checking baseline, and human baseline. We use ExactMatch, and a numeracy-focused (macro-averaged) F1 score as evaluation metrics to compare model performance which has been used in prior works \cite{dua2019drop}.

\paragraph{Heuristic Baselines along with Type Oracle:}
We assume that there is a type oracle that knows the question type. We add this to our heuristic baseline, since use of a single baseline across all eight types is not appropriate. In random baseline, we randomly select one of the options in case the question has multiple options (type 7 and type 3), a number between 0 to 100 for questions having a numerical answer and a random entity present in the passage for questions having a text segment from the passage as the answer. In the majority baseline, we select the most frequent answer for each type such as "Entailment" for NLI questions, 2nd option for questions having three options, most frequent number for questions having numerical answer and the major entity present in the passage for questions having span based answer.

\paragraph{Extended NumNet+v2}
We convert every type of question to RC format because RC has a passage component that represents local facts relevant to the questions. This allows easy integration of external knowledge (global facts) to the passage with no risk of getting it mixed with the question. This is useful because some of our data needs knowledge and this setting will help models in providing knowledge. We add a type classifier module to do this task. We design type classifier heuristically. Then, we convert each of the question types to RC format. For NLI questions, we use the premise sentence as passage, hypothesis as the question and append the string “Entailment, contradiction or neutral?” to the question so that it has a span based answer. For other questions, we tokenize the question string into its constituent sentences and use a heuristic approach to split the question string into passage and question. Furthermore, for option based questions, we append all the options at the end of the question. Figure \ref{motiv} illustrates the type conversion process. 

\paragraph{Bias-checking Baselines}
Recently, many works have shown that some of the popular NLP datasets have annotation biases which are exploited by language models \cite{poliak2018hypothesis}. In this baseline, we want to check if our dataset has any of those biases. We train our 
Extended NumNet+v2 model by heuristically removing the question completely and replacing it with the word "question". Similarly, we do the same by removing context. Context is different for different types such as passage in type 5 and type 6, premise in type 7, all sentences apart from question for type 8. Data which could not be split this way were converted manually.
\paragraph{Data Oversampling Baseline}
We try to tackle data imbalance by oversampling all types of data to the maximum size of all types.
\paragraph{Knowledge Hunting along with Extended NumNet+V2}
We create a cheat-sheet by accumulating all types of external knowledge which are needed to solve questions of various types. We use elasticsearch to hunt relevant knowledge sentences. We further filter them using a heuristic threshold of relevance. We append this knowledge in the beginning of the passage so that continuity is not broken between passage and question. Figure \ref{fig:knowledgemodel} illustrates our approach.
\paragraph{Human Baseline:}
Human baseline was calculated on 100 samples of each type (81 of Type 1) from the test set by averaging the scores of four individuals. 

\section{Results and Discussion}
Table \ref{results} shows the performance of various baseline models on our test set. Performance of all the baseline models is significantly less than human baseline. Answering questions in eight different setting using a single model is indeed a challenging task. 
We find that, in some of the cases, the model fails to distinguish among the question types. For example, it gave a span based answer where a number was expected and vice versa in some cases.
\paragraph{Dataset Bias:}
We performed multiple experiments to evaluate bias in our dataset. All bias-check baselines did not perform well even with the help of a type oracle.
This shows that our dataset has very less bias.

\paragraph{Which question types are hard to solve?}
Our results show that type 1 which requires numerical commonsense knowledge, is the hardest question type.
Similarly, Type 2, 4 and 8 appear to be comparatively harder from the rest. One pattern among these question types is that all of them expect the answer to be numeric. 
Numeric answer requires accurate calculation. So, models might have difficulty in learning the task with just data. This hypothesis is also justified from the drop in human performance for all these types.

\paragraph{Which question types are comparatively easier?}
Quantitative NLI has the best performance among all types. Also, performance on type 6 is better on type 5. Though both these datasets are inherited from the same parent. Models answer span based questions better as compared to numeric answers. Performance of model for type 3 questions further suggests that models find it easier to answer in an MCQ setting.

\paragraph{Do the knowledge retrieval help?}
Results show that knowledge help in improving performance of type 1, 2 and 4. It does not have significant difference for type 5, 6, 7, 8 which is as expected since questions of those types do not need external knowledge. This has been discussed in Section \ref{createdata}. However, seems like it acts as noise for type 3.
Conditional knowledge retrieval might help to mitigate the adverse effect.

\paragraph{Does oversampling help to overcome data imbalance?}
Even though oversampling results in higher performance in certain types, specifically the ones with smaller training data, it results in significant drop in performance in the other extreme, i.e types with bigger training data. Oversampling disproportionately might help to resolve this issue.

 \section{Conclusion}
 We have compiled a multifaceted dataset involving eight different types of data and define a task to solve this dataset in multi-task setting. We introduce numerical reasoning questions that require external knowledge, commonsense knowledge and domain knowledge. Based on results, we infer that performing well across all the eight tasks is more challenging than existing tasks. Also, the effect of our baseline performance on providing external knowledge to a language model promises the benefit of using cheat sheet for those tasks which need knowledge. We expect this dataset and the task to motivate researchers to analyze the generalization capability of neuro symbolic models. 
 Our future work will explore novel ways to do transfer learning in this multi task setting.


\bibliographystyle{named}
\bibliography{ijcai20}

\end{document}